\documentclass{article}
\usepackage[utf8]{inputenc}
\usepackage[margin = 1in]{geometry}
\usepackage{authblk}

\usepackage{amsmath, amssymb, amsthm, amsfonts, mathtools, multirow, mathrsfs, tabu, booktabs, lipsum, url, subcaption, graphicx,xcolor, tabu,  authblk, hyperref, blindtext}

\usepackage[english]{babel}
\usepackage[shortlabels]{enumitem}
 \usepackage[ruled,vlined, linesnumbered]{algorithm2e}
\usepackage[capbesideposition=outside,capbesidesep=quad]{floatrow}
\usepackage[framemethod=tikz]{mdframed}
\usepackage{bbm}

\allowdisplaybreaks

\newcommand{\appropto}{\mathrel{\vcenter{
  \offinterlineskip\halign{\hfil$##$\cr
    \propto\cr\noalign{\kern2pt}\sim\cr\noalign{\kern-2pt}}}}}
    
\newcommand{\R}{\mathbb{R}} 

\DeclareMathOperator*{\argmin}{argmin}

\title{Unsupervised detection of ash dieback disease (\emph{Hymenoscyphus fraxineus}) using diffusion-based hyperspectral image clustering}
\author[1]{Sam L. Polk}
\author[2]{Aland H. Y. Chan}
\author[3]{Kangning Cui}
\author[4]{Robert J. Plemmons}
\author[2]{David~A.~Coomes}
\author[1]{James M. Murphy\footnote{Corresponding Author: JM.Murphy@Tufts.edu\newline This work was partially funded by  the US National Science Foundation grants  NSF-DMS 1924513, NSF-CCF 1934553, and NSF-DMS 1912737.}}
\affil[1]{ Department of Mathematics, Tufts University}
\affil[2]{ Department of Plant Sciences, University of Cambridge}
\affil[3]{ Department of Mathematics, City University of Hong Kong}
\affil[4]{ Departments of Mathematics and Computer Science, Wake Forest University}
\date{}  
\begin{document}

\topmargin=0mm

\maketitle

\begin{abstract}

Ash dieback (\emph{Hymenoscyphus fraxineus}) is an introduced fungal disease that is causing the widespread death of ash trees across Europe. Remote sensing hyperspectral images encode rich structure that has been exploited for the detection of dieback disease in ash trees using supervised machine learning techniques. However, to understand the state of forest health at landscape-scale, accurate unsupervised approaches are needed. This article investigates the use of the unsupervised Diffusion and VCA-Assisted Image Segmentation (D-VIS) clustering algorithm for the detection of ash dieback disease in a forest site near Cambridge, United Kingdom. The unsupervised clustering presented in this work has high overlap with the supervised classification of previous work on this scene (overall accuracy = 71\%). Thus, unsupervised learning may be used for the remote detection of ash dieback disease without the need for expert labeling.

\end{abstract}

 
\section{Introduction}

Ash dieback disease, caused by the \emph{Hymenoscyphus fraxineus} ascomycete, poses a major threat to the health of European forests and ash-dependent biota~\cite{mckinney2014ash}. To aid epidemiologists and forest managers in the modeling and mitigation of ash dieback disease, healthy and infected trees must be identified at landscape-scale~\cite{chan2021monitoring}. Recent years have brought significant advances in methods for passive remote sensing of forest pathogens~\cite{stone2017application, waser2014evaluating}. For example, remote sensing hyperspectral images (HSIs) are high-dimensional images that encode rich structure that can be exploited by machine learning algorithms for the detection of diseases in forests~\cite{chan2021monitoring, stone2017application, waser2014evaluating}. While HSIs may be used for disease detection, the large volume of HSI data generated over forests makes manual labeling (generally required for supervised learning) infeasible. Thus, unsupervised machine learning algorithms are needed to produce accurate disease mappings of forests using HSIs.

This article implements the unsupervised Diffusion and VCA-Assisted Image Segmentation  (D-VIS) on an HSI generated over a forest in Madingley Village, near Cambridge,  United Kingdom~\cite{chan2021monitoring}. D-VIS is closely related to Diffusion and Volume maximization-based Image Clustering (D-VIC), which has been shown to successfully recover ground truth labels from benchmark HSIs~\cite{DVIS}. We compare an unsupervised D-VIS clustering to a disease mapping generated by a supervised Random Forest (RF)~\cite{chan2021monitoring}. High levels of overlap were observed between unsupervised and supervised partitions, indicating that unsupervised learning---and D-VIS, in particular---may be used to detect ash dieback disease in European forests even when no ground truth labels are available. 

The rest of this article is structured as follows. In Section \ref{sec: background}, background material is provided on HSI segmentation, diffusion geometry, spectral unmixing, and the D-VIS clustering algorithm. In Section \ref{sec: numerical experiments}, the Madingley HSI is described, and disease mappings obtained by D-VIS are presented. We conclude and suggest directions for future work in Section \ref{sec: conclusion}.  

\section{Background} \label{sec: background}

\subsection{Hyperspectral Image Segmentation} \label{sec: HSIS}

Let $X = \{x_i\}_{i=1}^n\subset~\R^D$ be the set of HSI pixel spectra, where $n$ and $D$ indicate the number of pixels and bands, respectively. HSI segmentation algorithms partition an HSI into \emph{clusters} of pixels $\{X_k\}_{k=1}^K$~\cite{eismann2012hyperspectral, friedman2001elements}. Ideally, any two pixels in the same cluster will be similar, but two pixels from different clusters will be dissimilar~\cite{friedman2001elements}.
\emph{Supervised} HSI segmentation algorithms rely on some fraction of the image's ground truth labels to obtain a partition of $X$. Conversely, \emph{unsupervised} HSI segmentation algorithms (also called HSI clustering algorithms) require no ground truth labels to partition $X$~\cite{friedman2001elements}. 

\subsection{Diffusion Geometry} \label{sec: diffusion geometry}

Graph-based HSI clustering algorithms treat each pixel as a node in an undirected graph~\cite{coifman2006diffusion}. The edges between pixels can be encoded in an adjacency matrix $\textbf{W}\in\R^{n\times n}$, where $\textbf{W}_{ij}=1$ if $x_j$ is one of the $N$ $\ell^2$-nearest neighbors of $x_i$ and $\textbf{W}_{ij}=0$ otherwise. Define $\textbf{P} = \textbf{D}^{-1}\textbf{W}$, where $\textbf{D}\in\R^{n\times n}$ is the diagonal \emph{degree matrix} defined by $\textbf{D}_{ii} = \sum_{j=1}^n \textbf{W}_{ij}$. We identify  $\textbf{P}$ as the transition matrix for a Markov diffusion process on HSI pixels. Assuming $\textbf{P}$ is irreducible and aperiodic, there is a unique $\pi\in\R^{1\times n}$ satisfying $\pi\textbf{P}=\pi$~\cite{coifman2006diffusion}.  

\emph{Diffusion distances} enable the comparison of HSI pixels in the context of the diffusion process encoded in $\textbf{P}$~\cite{ coifman2006diffusion, maggioni2019learning}. Define 
\[D_t(x_i, x_j) = \sqrt{\sum_{k=1}^n \frac{\left(\left(\textbf{P}^t\right)_{ik}-\left(\textbf{P}^t\right)_{jk}\right)^2}{\pi_k} }\] 
to be the diffusion distance between $x_i$ and $x_j$ at time $t\geq 0$~\cite{coifman2006diffusion}. For datasets with highly coherent and well-separated latent cluster structure, the maximum within-cluster diffusion distance is bounded away from the minimum between-cluster diffusion distance across a large range of $t$~\cite{maggioni2019learning, murphy2021multiscale}. As such, diffusion distances can be used to learn latent structure in HSIs.

\subsection{Spectral Unmixing} \label{sec: spectral unmixing} 

HSIs are usually generated at a relatively coarse spatial resolution (often with a spatial resolution as high as 10~m)~\cite{eismann2012hyperspectral}. As such, in many scenarios (e.g., disease mapping in forests), a single pixel may correspond to a spatial region containing multiple materials. If $m$ is the number of materials present in the scene, the goal of a linear \emph{spectral unmixing} algorithm is to find two nonnegative matrices, $\textbf{A}\in\R^{n\times m}$ and $\textbf{U} =~(u_1, u_2, \dots, u_m)^\top\in \R^{m\times D}$  such that $x_i \approx \sum_{j=1}^m\textbf{A}_{ij}u_j$ for each $x_i\in X$~\cite{chan2011simplex, nascimento2005VCA}. Ideally, the rows of $\textbf{U}$ will encode  the intrinsic spectral signatures associated with materials in the scene, while the rows of $\textbf{A}$ encode the relative frequency that those materials appear in a given pixel.  Information about material abundance can be summarized using \emph{pixel purity}, defined for each $x_i\in X$ by 
\[\eta(x_i) = \frac{\max_{1\leq j \leq m} \textbf{A}_{ij}}{\sum_{j=1}^m \textbf{A}_{ij}}.\] 
The function $\eta(x)$ will be nearly 1 if the spatial region corresponding to the pixel $x$ contains predominantly one material, and $\eta(x)$ will be small for mixed pixels containing many materials~\cite{ADVIS, DVIS}.

\subsection{Diffusion and VCA-Assisted Image Segmentation} \label{sec: D-VIS}

In its first stage, D-VIS (Algorithm \ref{alg: D-VIS}) locates $K$ pixels that exemplify underlying structure in the HSI. To locate these \emph{cluster modes}, D-VIS first calculates  $\eta(x)$ using HySime~\cite{bioucas2008HySime} to learn $m$ and VCA~\cite{bro1997fast, nascimento2005VCA} to learn $\textbf{A}$ and $\textbf{U}$. This differs slightly from D-VIC, which relies on Alternating Volume Maximization to learn $\mathbf{U}$~\cite{chan2011simplex,DVIS}. Next, D-VIS calculates density: 
\[p(x) = \sum_{y\in NN(x,N)}\exp\left(-\frac{\|x-y\|_2^2}{\sigma_0^2}\right),\] where $NN(x,N)$ is the set of $N$ $\ell^2$-nearest neighbors of $x$ in $X$ and $\sigma_0>0$ is a \emph{density scale} that controls interactions between points. The functions $p(x)$ and $\eta(x)$ are combined into a single measure 
\[\zeta(x) = \frac{2\bar{p}(x)\bar{\eta}(x)}{\bar{p}(x) + \bar{\eta}(x)},\]
where $\bar{p}(x) = \frac{p(x)}{\max_{y\in X}p(y)}$ and $\bar{\eta}(x) = \frac{\eta(x)}{\max_{y\in X}\eta(y)}$. As the harmonic mean of pixel purity and density, $\zeta(x)$ will be large if and only if $x$ is both modal (i.e.,  close to its $N$ $\ell^2$-nearest neighbors) and representative of a single material in the scene (i.e.,  high-purity).

Cluster modes are identified in D-VIS as the maximizers of $\mathcal{D}_t(x) = \zeta(x) d_t(x)$, where
\begin{align*}
    d_t(x) = \begin{cases}
    \min\limits_{y\in X}\left\{D_t(x,y)| \zeta(y)\geq \zeta(x)\right\} & x = \argmin\limits_{y\in X} \zeta(y),\\
    \max\limits_{y\in X} D_t(x,y) & \text{otherwise.}
    \end{cases}
\end{align*}
The function $d_t(x)$ outputs the diffusion distance at time $t$ between $x$ and its diffusion distance-nearest neighbor of higher $\zeta$-value. Thus, $\mathcal{D}_t$-maximizers are high-density, high-purity pixels that are far in diffusion distance at time $t$ from other high-density, high-purity pixels. As such, these $K$ points are suitable exemplars for the materials in the scene. Cluster modes are given unique labels, and each non-modal $x\in X$ is, in order of non-increasing $\zeta(x)$, given the label of its $D_t$-nearest neighbor of higher $\zeta$-value that is already labeled.  

\begin{algorithm}[t]
\caption{Diffusion and VCA-Assisted Image Segmentation (D-VIS) \label{alg: D-VIS}}
\KwIn{$X$ (HSI), $K$~(\#~clusters), $N$ (\# neighbors), $\sigma_0$ (density scale), $t$ (diffusion time)}
\KwOut{$\hat{\mathcal{C}}= \{X_k\}_{k=1}^K$ (clustering)}
Calculate $\eta(x)$ by implementing HySime~\cite{bioucas2008HySime} to estimate $m$ and VCA~\cite{bro1997fast, nascimento2005VCA} to estimate $\textbf{U}$ and $\textbf{A}$\;  
Calculate $\zeta(x)$ using density $p(x)$ (see Section \ref{sec: D-VIS})\; 
Label cluster modes $\hat{\mathcal{C}}(x_{m_k}) = k$ for $1\leq k \leq K$, where $\{x_{m_k}\}_{k=1}^K$ are the $K$ maximizers of  $\mathcal{D}_t(x) = \zeta(x) d_t(x)$; $d_t(x)$ is as in Section \ref{sec: D-VIS}\; 
In order of non-increasing $\zeta(x)$, assign unlabeled points $x\in X$ the label $\hat{\mathcal{C}}(x) = \hat{\mathcal{C}}(x^*)$, where $x^* =\argmin\limits_{y\in X}\left\{D_t(x,y) \Big|\; \zeta(y)\geq \zeta(x)\text{ and } \hat{\mathcal{C}}(y)>0\right\}$\; 
\end{algorithm}
  
\section{Unsupervised Ash Dieback Detection} \label{sec: numerical experiments}

\subsection{Madingley Hyperspectral Image}

This article presents the implementation of D-VIS on hyperspectral  data collected by a human-crewed aircraft  over a $512 \text{m} \times 356 \text{m}$ region of temperate deciduous forest near Madingley, on the outskirts of Cambridge, United Kingdom, in August 2018~\cite{chan2021monitoring}. Spectral reflectance was recorded using a Norsk Elektro Optikk hyperspectral camera (Hyspex VNIR 1800) at a spectral resolution of 3.26 nm across wavelengths 410-1001 nm and at a high spatial resolution of 0.32 m. Thus, reflectance at a total of $D=186$ spectral bands was recorded for $1816835$ pixels in a scene with spatial dimensions of $1601\times 1113$. QUick Atmospheric Correction~\cite{eismann2012hyperspectral} was implemented to remove atmospheric effects on pixel spectra, and spectral signatures were normalized to prevent differences in illumination from corrupting classification results~\cite{chan2021monitoring}. 

\subsection{Supervised Classification of Ash Dieback Disease} \label{sec: Supervised}

\begin{figure}[t]
    \centering
    \includegraphics[width = 0.35\textwidth]{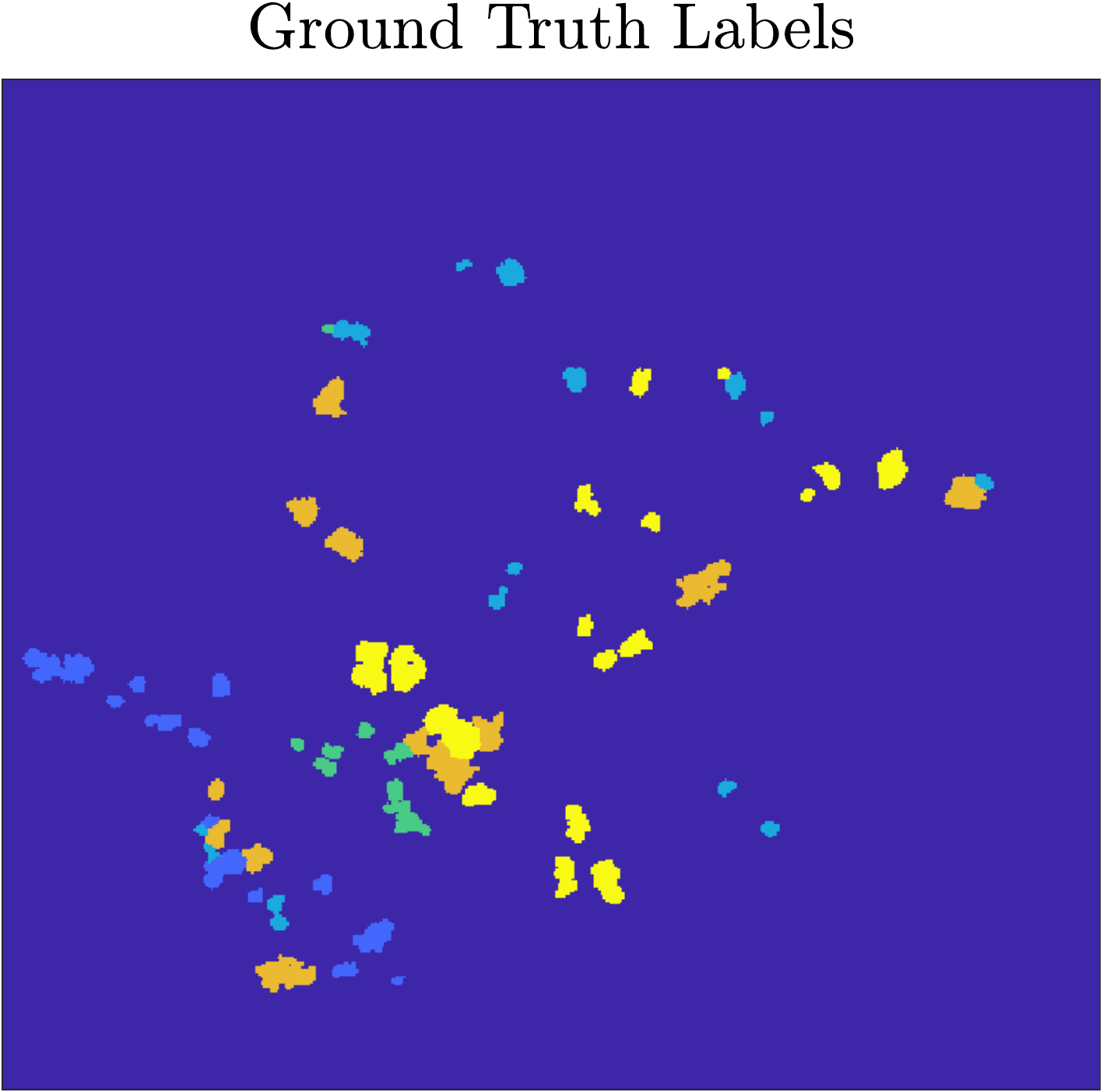} \hspace{0.2in}
    \includegraphics[width = 0.35\textwidth]{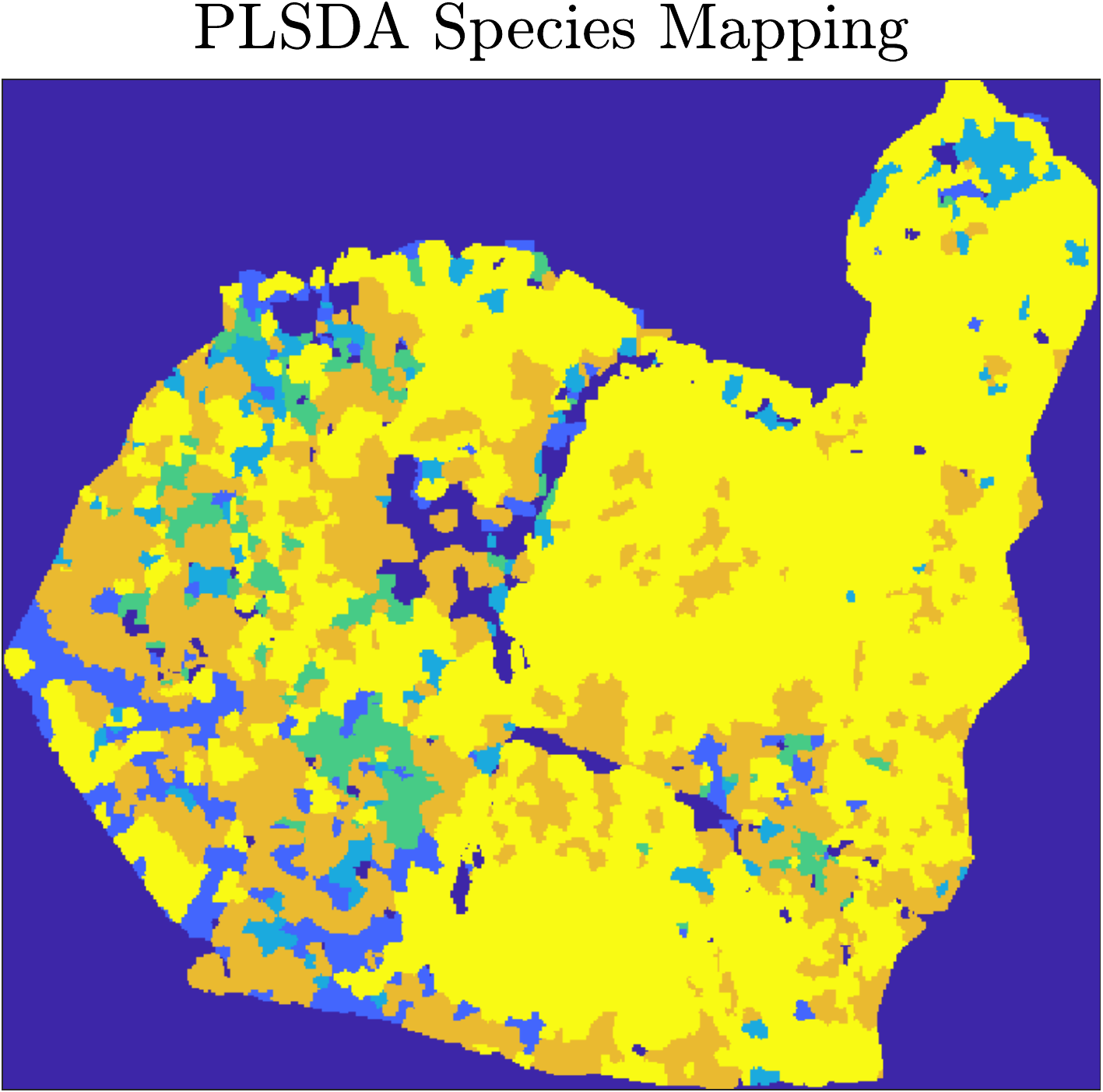}
    \caption{Ground truth labels and PLSDA species mapping for the Madingley HSI~\cite{chan2021monitoring}. Colors indicate different tree species. The class marked in yellow corresponds to ash trees. }
    \label{fig:GT Data}
\end{figure}

To enable the detection of ash dieback disease in the Madingley scene, a Partial Least Squares Discriminant Analysis (PLSDA) classifier was trained to locate pixels corresponding to ash trees in the Madingley scene~\cite{chan2021monitoring}. Ground truth labels were collected for 166 tree crowns in the Madingley region as well as 256 tree crowns from three other forest sites near Cambridge~\cite{chan2021monitoring}. Manually delineated and labeled tree crowns from these four forest sites were separated into training (70\%) and testing (30\%) sets. The resulting species mapping was highly accurate, with an overall accuracy of 85.3\% on the testing dataset~\cite{chan2021monitoring}. The ground truth labels and PLSDA species mapping are visualized in Fig. \ref{fig:GT Data}.  

Ash dieback disease was identified using a supervised RF classifier~\cite{chan2021monitoring}. The RF was trained to detect three disease classes---infected, severely infected, and healthy---among trees labeled ash in the PLSDA species mapping. Models were trained and tested using the average pixel spectra of tree crowns, where the testing set included 16 trees from each disease class. The resulting RF classifier demonstrated strong recovery of ash dieback disease in the Madingley scene, with an overall accuracy of 77.1\%~\cite{chan2021monitoring}. 

\subsection{Unsupervised Classification of Ash Dieback Disease}\label{sec: unsupervised}

This section presents the results of unsupervised clustering of the Madingley HSI using the D-VIS clustering algorithm. Because ash trees affected by dieback often have a mosaic of healthy and dead branches, the pixels corresponding to visibly infected trees at the original 0.32 m spatial resolution may resemble the pixels of healthy trees~\cite{chan2021monitoring}. Therefore, to provide a more holistic assessment of tree health (as opposed to that of the health of individual branches), the HSI was downsampled to a spatial resolution of 1.28 m using bicubic interpolation~\cite{keys1981bicubic} so that each pixel covered multiple branches.  

Pixels corresponding to species other than ash in the PLSDA species mapping were discarded, leaving $n=72775$ pixels across a $401\times 279$ scene. D-VIS relied on a sparse $K$-nearest neighbors graph with $N=150$ edges per pixel, a density scale $\sigma_0 = 3.89\times10^{-4}$, and $t=2^5$. We set $K=2$ so that classes corresponded to healthy and dieback-infected trees. After cluster analysis, majority voting was implemented among pixels in each delineated tree crown, yielding an unsupervised tree crown-level disease mapping.

\begin{figure}[t]
    \centering
    \includegraphics[width = 0.35\textwidth]{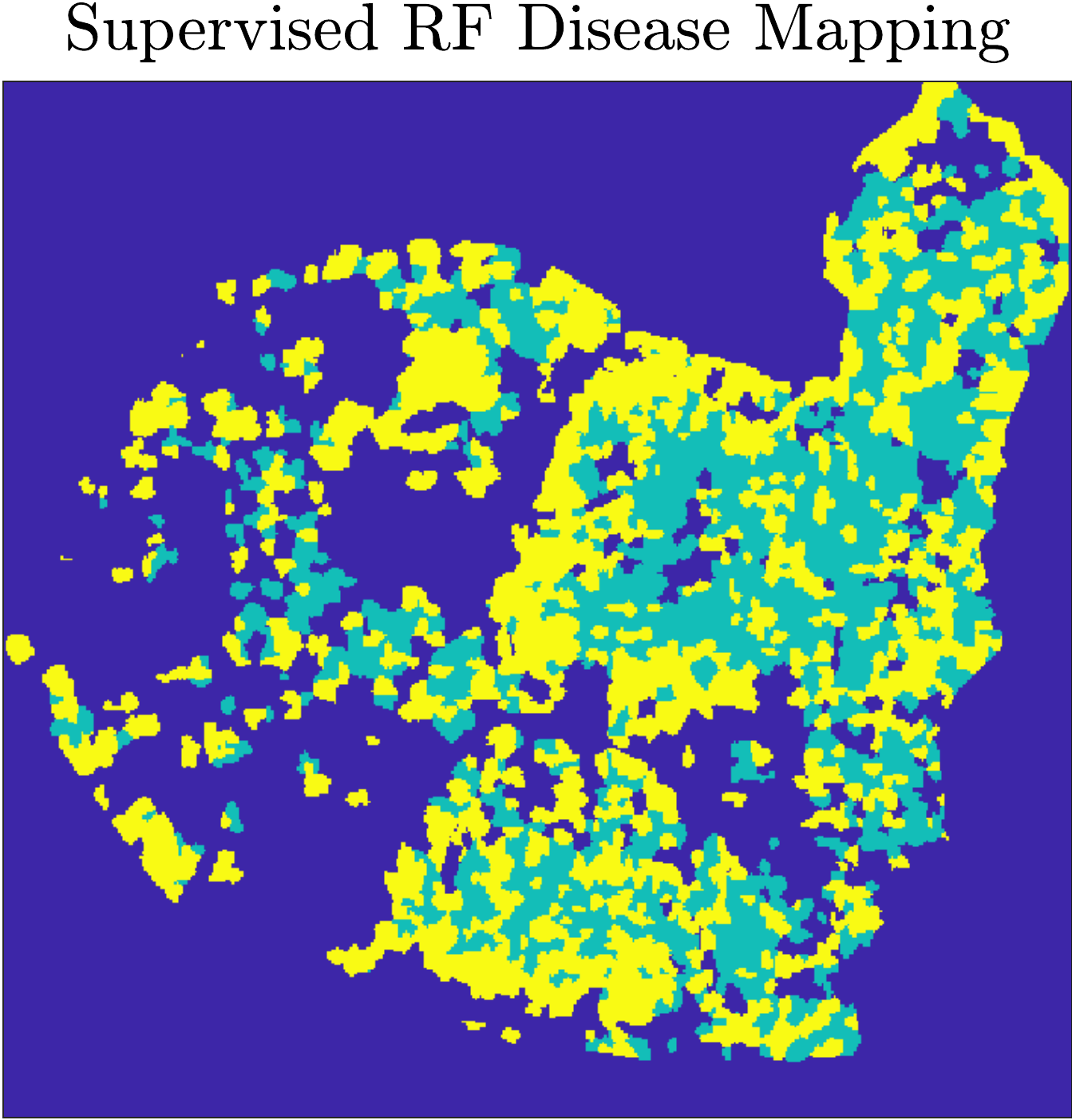} \hspace{0.2in}
    \includegraphics[width = 0.35\textwidth]{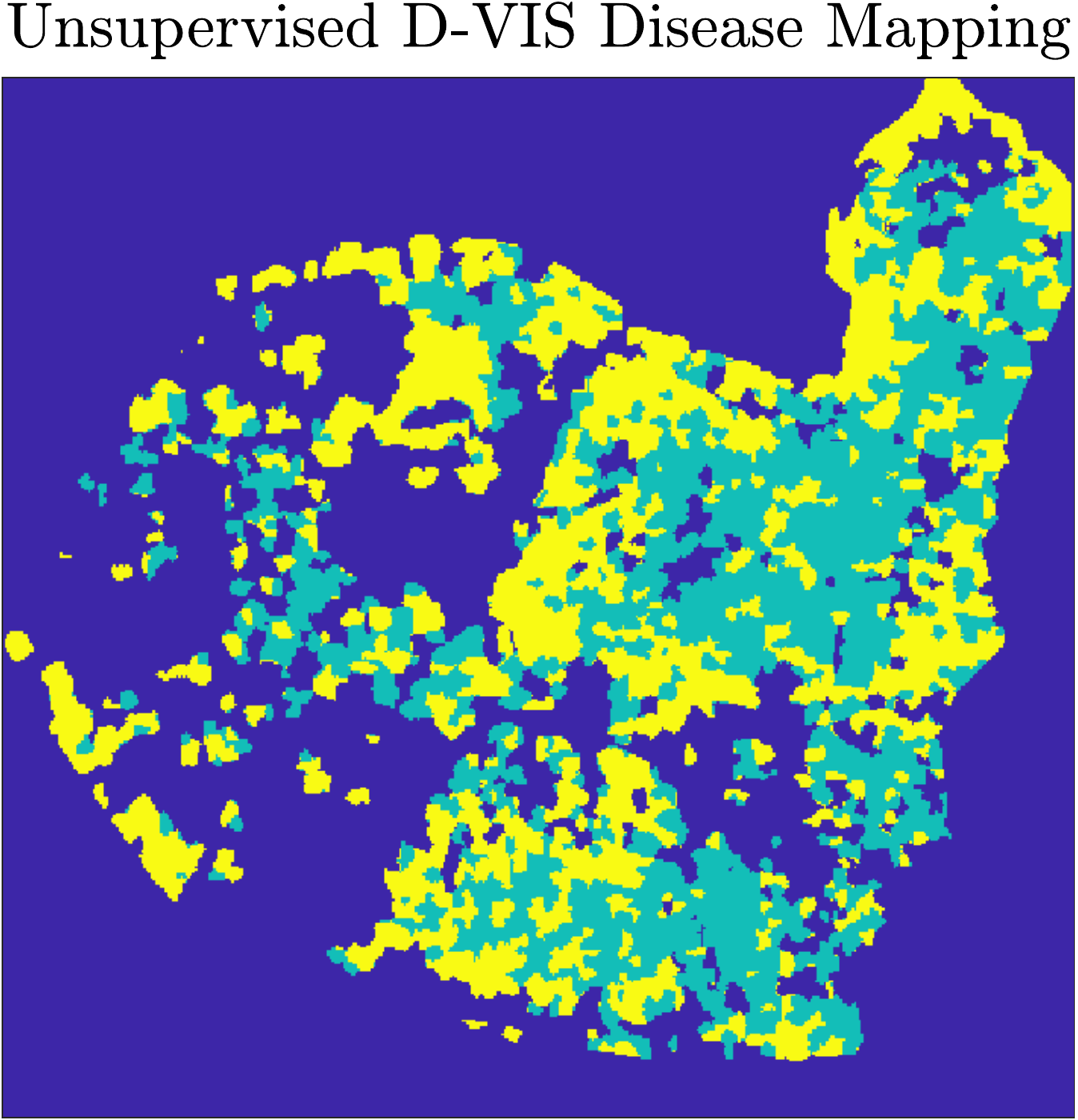} 

    \caption{Supervised RF~\cite{chan2021monitoring} and unsupervised D-VIS disease mappings of the Madingley HSI. Significant overlap exists between the two labelings, indicating that unsupervised methods such as D-VIS may be used for ash dieback disease mapping when no ground truth labels are available.}
    \label{fig: disease}
\end{figure}

For validation, the unsupervised D-VIS clustering was compared against the supervised RF classification of ash dieback disease obtained in prior work~\cite{chan2021monitoring} after aligning labels using the Hungarian algorithm. The two dieback classes in the RF disease mapping  (infected and severely infected) were combined, yielding a single ``dieback'' class. A matching matrix summarizing the overlap between the supervised RF and unsupervised D-VIS labelings is provided in Table \ref{tab: matching matrix}, while disease mappings are visualized in Fig. \ref{fig: disease}. D-VIS and RF disease mappings exhibited a high level of overlap, with D-VIS achieving an overall accuracy of 71.0\% and average accuracy of 71.3\%. Thus, unsupervised clustering algorithms such as D-VIS may be used for the detection of ash dieback disease using hyperspectral data, even when no ground truth labels are available.

\begin{table}[b]
    \centering
    \begin{tabular}{|c|cc|c|}
    \hline
        & \textbf{Healthy}       & \textbf{Dieback} & \textbf{Producer's Acc.}  \\ \hline 
       \textbf{Healthy}          &  \textbf{27460} & 12895 & 68.0\% \\
       \textbf{Dieback} & 8238 & \textbf{24182} & 74.6 \%\\  \hline 
       \textbf{User's Acc.}  &  76.9\%  & 65.2\% &  \\ \hline 
    \end{tabular}
    \caption{Matching matrix showing overlap between the unsupervised D-VIS and supervised RF ash dieback mappings~\cite{chan2021monitoring}. Rows summarize how pixels labeled by the RF in a fixed class were classified by D-VIS. ``Acc.'' indicates Accuracy.  }
    \label{tab: matching matrix}
\end{table}

\section{Conclusions} \label{sec: conclusion}

We conclude that the unsupervised D-VIS clustering algorithm can successfully identify ash dieback disease from remote sensing HSIs. Future work includes implementing the pipeline developed in this article on additional forest regions as further validation, as well as scaling up our approach to be implemented on larger forests.  Moreover, we expect the unsupervised learning procedure outlined in this article to be useful for the remote detection of other damaging agents in forests and hope to consider this problem in future work. Finally, the performance of D-VIS is likely to improve upon further modification (e.g., modifying for active learning~\cite{ maggioni2019LAND, murphy2020spatially, ADVIS} or adding spatial regularization~\cite{murphy2020spatially, murphy2019spectral, polk2021multiscale}).

\section{Acknowledgments}

We acknowledge 2Excel Geo (in particular, Dr. Chloe Barnes) for collecting the high-resolution HSI used in this study. We thank the University of Cambridge for critical support and access to the Madingley field site. We thank the Wildlife Trust for Bedfordshire, Cambridgeshire \& Nottinghamshire for allowing access to other woodland sites. Finally, we thank Dr. Carola-Bibiane Sch{\"o}nlieb for operating the INTEGRAL research group during the Zoom meetings of which many discussions that motivated this work occurred.

\bibliographystyle{siam} 
\bibliography{ref}

\end{document}